\documentclass[conference]{IEEEtran}
\usepackage{cite}
\usepackage{amsmath,amssymb,amsfonts}
\usepackage{algorithmic}
\usepackage{graphicx}
\usepackage{textcomp}
\usepackage{xcolor}
\usepackage{booktabs}
\usepackage{multirow}
\usepackage{capt-of}
\usepackage{bbm}
\usepackage{hyperref}
\usepackage[normalem]{ulem}
\usepackage[inkscapelatex=false]{svg}
\usepackage{makecell}
\usepackage{float}

\hypersetup{
    colorlinks=true,
    linkcolor=black,
    citecolor=black,
    urlcolor=cyan,
    filecolor=black
}

\definecolor{linkblue}{RGB}{0,112,193}
\newcommand{\link}[1]{\textcolor{linkblue}{#1}}

\def\BibTeX{{\rm B\kern-.05em{\sc i\kern-.025em b}\kern-.08em
    T\kern-.1667em\lower.7ex\hbox{E}\kern-.125emX}}
\begin{document}

\title{Learning-Guided Sparsification of Dynamic Graphs in Robotic Exploration
% \\ \thanks{Source code is available at \textcolor{cyan}{\uline{\href{https://github.com/avs-origami/graphsparse}{https://github.com/avs-origami/graphsparse}}}}
}

\author{\IEEEauthorblockN{1\textsuperscript{st} Adithya V. Sastry}
\IEEEauthorblockA{
% \textit{dept. name of organization (of Aff.)} \\
\textit{Independent Researcher}\\
Knoxville, TN, USA \\
avs.origami@gmail.com
}
\and
\IEEEauthorblockN{2\textsuperscript{nd} Bibek Poudel}
\IEEEauthorblockA{
% \textit{Min H. Kao Department of Electrical Engineering and Computer Science} \\
\textit{Purdue University}\\
West Lafayette, IN, USA \\
poudel16@purdue.edu}
\and
\IEEEauthorblockN{3\textsuperscript{rd} Weizi Li}
\IEEEauthorblockA{
% \textit{Department of Computer Science and Engineering} \\
\textit{University of California, Riverside}\\
Riverside, CA, USA \\
weizi.li@ucr.edu}
}

\maketitle

\begin{abstract}
Many robotic exploration algorithms rely on graph structures for frontier-based exploration and dynamic path planning. However, these graphs grow rapidly, accumulating redundant information and impacting performance. We present a hybrid transformer-based framework trained with Proximal Policy Optimization which complements exploration algorithms by pruning these graphs during exploration, limiting their growth and reducing the accumulation of excess information. The framework was evaluated on simulations of a robotic agent using Rapidly-Exploring Random Trees to carry out frontier-based exploration, where the learned policy reduces graph size by up to 96\%. We find preliminary evidence that our framework enables effective, generalizable exploration under reduced information density, consistently outperforming the randomly pruned baseline and improving both exploration efficiency and computational efficiency beyond the unpruned baseline for complex long-horizon exploration tasks. To the best of our knowledge, these results are the first suggesting the viability of RL to prune dynamic graphs used in robotic exploration algorithms.

\end{abstract}


\section{Introduction}
\label{sec:intro}

Autonomous systems have achieved remarkable progress in navigating complex, unstructured environments, enabling robots to carry out tasks that were previously limited to human operators~\cite{wang2023autonomous}. Search and rescue, space exploration, and surveillance all stand to benefit from autonomous robotic exploration, as these tasks often require operation in environments too hazardous or inaccessible for humans~\cite{azpurua2023survey}. However, many exploration algorithms rely on graph or tree structures for path planning and waypoint creation~\cite{dang2020subterranean, patil2023multirobotsearch}, and these structures grow continuously as the agent explores, accumulating redundant information that can degrade performance.

By pruning these graphs, we can allow a robot to focus on pushing frontiers with greater potential for information gain, reducing the overhead of maintaining large graph structures. Yet, the relationships between the exploration graph and the environment shift as the robot gains new information about its surroundings~\cite{dang2020subterranean, patil2023multirobotsearch, umari2017rrt}. Pruning decisions must therefore be made sequentially under partial observability, with outcomes that are delayed and difficult to attribute to individual actions. In the absence of clear heuristic signals, reinforcement learning (RL) is a promising tool for learning an effective pruning strategy, as it is designed to handle sequential decision-making with sparse, delayed feedback, without supervised labels.

\begin{figure}[t!]
    \centering
    \includegraphics[width=0.29\linewidth]{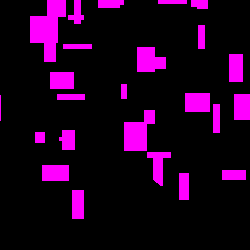}
    \hfill
    \includegraphics[width=0.29\linewidth]{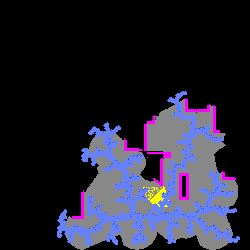}
    \hfill
    \includegraphics[width=0.29\linewidth]{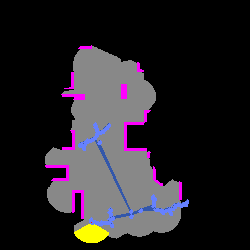}
    \caption{\textbf{Left:} An example environment the robot explores. \textbf{Middle:} Exploration progress without pruning. \textbf{Right:} Exploration progress with pruning over a comparable time period. Magenta blocks are obstacles, gray area represents explored free space, black area represents unexplored free space, blue represents the exploration graph, and yellow represents the robot's field of view. Pruning reduces the exploration graph size by up to $96$\%, demonstrating the potential for intelligent sparsification in graph-based exploration.}
    \label{fig:simulator}
    \vspace{-10pt}
\end{figure}

We introduce a hybrid transformer-based framework that complements traditional frontier-based robotic exploration by utilizing RL to prune the dynamic exploration graphs these algorithms rely upon, reducing the size of these graphs by up to 96\% (Fig.~\ref{fig:simulator}). Central to our framework is a spatial parameterization of the environment which enables the learned policy to act upon the exploration graph while remaining otherwise decoupled from frontier selection and path planning. Preliminary evidence shows that our framework enables effective, generalizable exploration under reduced information density, consistently outperforming the randomly pruned baseline. In complex long-horizon exploration tasks, our framework improves exploration efficiency by up to 2.1 percentage points and improves computational efficiency by up to $2\times$ relative to the unpruned baseline. To the best of our knowledge, these results are the first to suggest the viability of RL for intelligent sparsification of dynamic graphs in graph-based robotic exploration algorithms. Source code can be found in the GitHub repository\footnote{\link{\textbf{\underline{\url{https://github.com/avs-origami/graphsparse}}}}}.

\section{Related Work}
\label{sec:related}

\begin{figure*}[t!]
    \centering
    \includegraphics[width=0.85\linewidth]{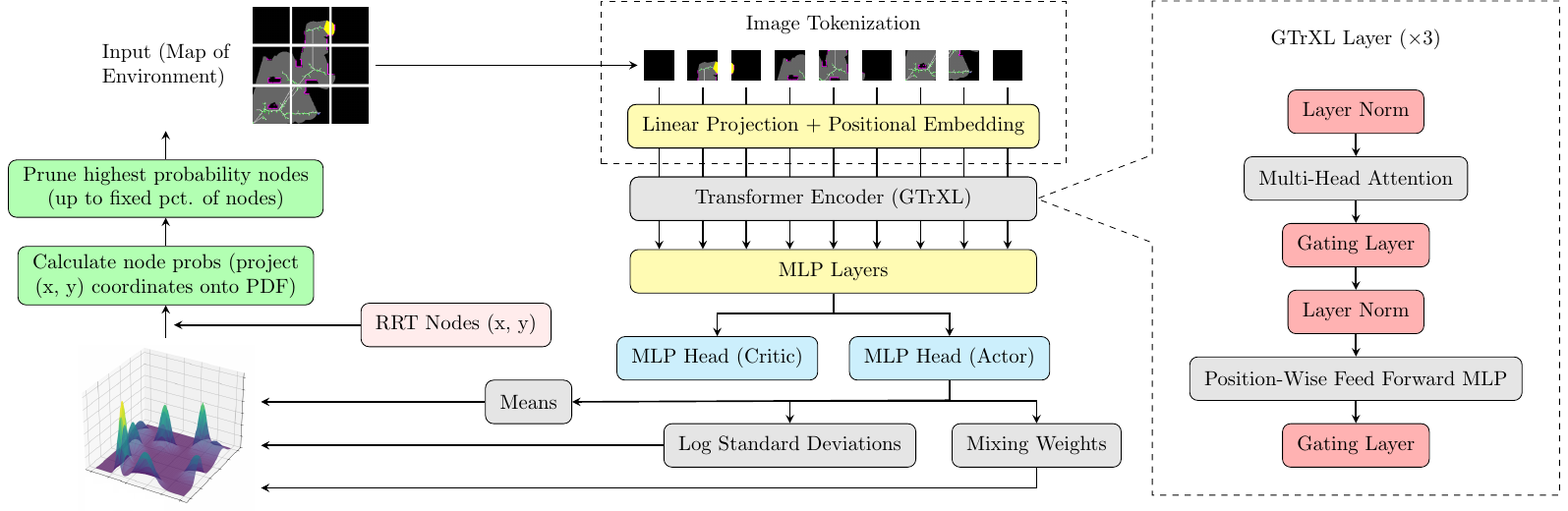} 
    \vspace{-205pt}
    \caption{Overview of the framework for pruning robotic exploration graphs. An image of the current map (top left) is split into patches, tokenized, and positionally embedded before passing through a Gated Transformer-XL encoder and multi-layer perceptron layers for feature extraction. The extracted features feed into separate actor and critic heads, where the actor predicts the means, standard deviations, and mixing weights that define a Gaussian Mixture Model (GMM) over the environment. To determine which nodes to prune, the coordinates of nodes in the exploration graph are projected onto this GMM. The environment is then split into the same patches created by the tokenizer, and the highest probability nodes in each patch are retained; all others are pruned.}
    \label{fig:modelarch}
    \vspace{-10pt}
\end{figure*}

Frontier-based exploration~\cite{yamauchi1997frontier} is a widely used paradigm in which graph or tree structures are used to identify frontiers (the boundary between explored and unexplored regions) and guide pathfinding towards these regions, continuously maximizing information gain~\cite{umari2017rrt, hyejeong2020graph, dang2020subterranean}. Deep reinforcement learning (DRL) also sees extensive use in robotic exploration~\cite{liu2023learning}, with methods ranging from using direct policy learning to identify exploration targets and carry out exploration~\cite{luperto2021exploration, schmid2022fast}, to using graph neural networks to make navigation decisions based on specialized graph representations of the environment~\cite{chen2020autonomous, chen2021zeroshot, luo2019multi, zhang2022heirarchical}. Canonical algorithms may be better suited to rapid replanning or dynamic environments, while DRL approaches can outperform them in more structured settings~\cite{arce2023comparison}. However, purely learned approaches must relearn basic exploration behaviors from scratch and can struggle when encountering diverse environments. Rather than using a learned policy directly for exploration, we combine a canonical exploration algorithm with DRL for the subtask of pruning the exploration graph. This preserves the algorithmic robustness of frontier-based methods while using RL to learn the complex, evolving relationship between the graph and the environment, thereby enabling exploration with reduced information.

The pruning task requires the model to learn how sequences of decisions affect future exploration over long horizons. To capture these long-term dependencies, we rely on the Gated Transformer-XL (GTrXL)~\cite{parisotto2020stabilizing}. Transformers~\cite{vaswani2017attention} are effective for feature representation across a range of tasks~\cite{devlin2018bert, radford2019langauge, dosovitskiy2020vit}, and recent works extend their capacity for long-term memory~\cite{dai1901transformer, ni2024transformers}. GTrXL adds gating mechanisms that stabilize training under the non-stationary data distributions typical in RL, making it well-suited for our task where the state and action spaces shift as exploration progresses.

In short, while existing studies address robotic exploration through either purely algorithmic strategies~\cite{umari2017rrt, patil2023multirobotsearch, hyejeong2020graph, yamauchi1997frontier, dang2020subterranean} or fully learned approaches~\cite{liu2023learning, luperto2021exploration, schmid2022fast, chen2020autonomous, chen2021zeroshot, luo2019multi}, we introduce a hybrid approach that leverages RL to perform intelligent sparsification on the exploration graph. By interfacing the algorithmic and learned components through spatial parameterization of the environment, we preserve the robustness, adaptability, and interpretability of graph-based exploration while leveraging learned policies to reduce redundant information.
\section{Methodology}
\label{sec:methodology}

We define the task as pruning nodes from the exploration graph of a frontier-based robotic exploration algorithm, to produce a sparse graph that is at least as effective as the original. To simplify the problem, we assume a fixed, known environment size, which enables fixed-size model inputs and outputs. Our solution framework is outlined in Fig.~\ref{fig:modelarch}.

\subsection{Markov Decision Process}

We formulate the graph pruning strategy as a partially observable Markov decision process $(S, A, T, R, \Omega, O, \gamma)$, where $S$ is the set of environment states, $A$ is the set of actions, $T: S \times A \times S \to [0, 1]$ is the probabilistic state transition function, and $R: S \times A \to \mathbb{R}$ is the reward function. Due to partial observability, $\Omega$ is the set of observations, $O: S \times A \times \Omega \to [0, 1]$ is the observation probability function, and $\gamma \in [0, 1)$ is the discount factor. At each timestep, the agent observes an incomplete view of the environment and selects a pruning action to maximize the cumulative discounted reward $G_t=\sum_{k=t}^T{\gamma^{k-t}r_k}$. We train the policy with Proximal Policy Optimization (PPO)~\cite{schulman2017proximal} and Generalized Advantage Estimation~\cite{schulman2015high}, as PPO is effective for problems with delayed or sparse reward signals~\cite{burda2018exploration}.

\begin{figure*}[t!]
    \centering
    \includegraphics[width=0.9\linewidth]{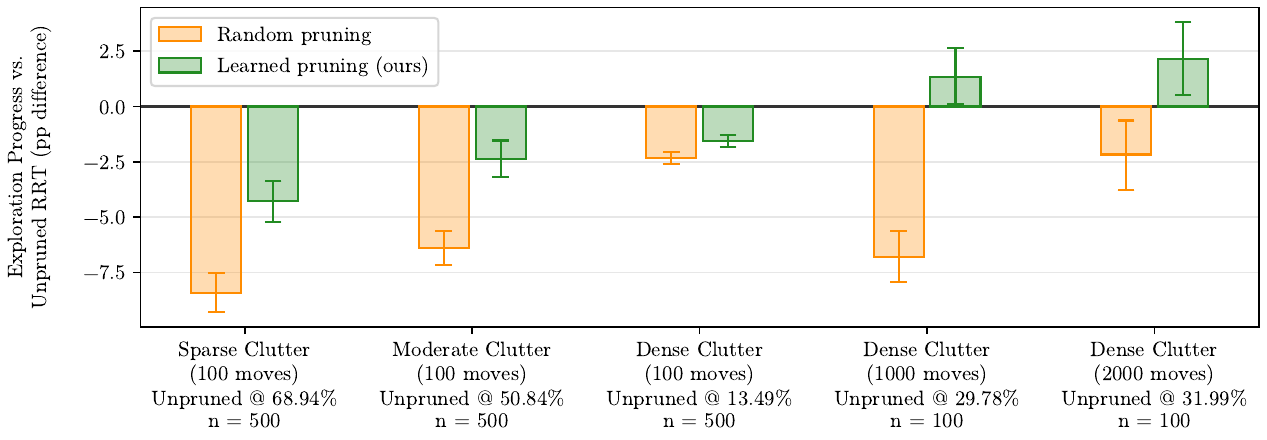}
    % \includegraphics[width=0.3\linewidth]{Figures/imagem_convertida(5).png}
    % \hspace{0.005\linewidth}
    % \includegraphics[width=0.3\linewidth]{Figures/imagem_convertida(4).png}
    % \hspace{0.005\linewidth}
    % \includegraphics[width=0.3\linewidth]{Figures/imagem_convertida(3).png}
    \vspace{-8pt}
    \caption{Exploration progress across various environment complexities and time horizons, measured as a percentage-point (pp) difference from the baseline (exploration without pruning). Learned pruning (ours) remains within $4.5$~pp of the exploration progress without pruning and consistently exceeds the exploration progress under uniform random pruning. Further, learned pruning generalizes to a variety of exploration tasks despite being trained on only high-complexity short-horizon simulations. As both the complexity of environments and the exploration time horizon increase, learned pruning becomes increasingly advantageous. Long-horizon runs use $n=100$ instead of the $n=500$ used for short-horizon runs due to increased computational cost of running long-horizon experiments. Error bars are roughly $\pm1$ SE of the difference in mean exploration progress. Note that the stated number of "moves" for each experiment refers to the value set for both the maximum number of robot moves ($\lambda_M$) and the maximum number of RRT growth attempts ($\lambda_R$).}
    \label{fig:reward2}
    \vspace{-12pt}
\end{figure*}

\vspace{4pt}
\noindent \textbf{State.} In contrast to approaches which rely on specialized graph representations of the environment~\cite{chen2020autonomous, chen2021zeroshot, zhang2022heirarchical, luo2019multi}, the state is an image of the current environment map with the exploration graph as an overlay, eliminating additional processing to encode spatial relationships between the exploration graph and the environment. The image is tokenized following the Vision Transformer~\cite{dosovitskiy2020vit} where it is split into fixed-size patches, linearly embedded, and combined with positional embeddings. Self-attention then integrates information from across the whole image, unlike other methods like convolutional neural networks, which consider only local context~\cite{dosovitskiy2020vit}.

\vspace{4pt}
\noindent \label{sec:methodology:mdp:action}\textbf{Action.} Since the set of nodes in the exploration graph is constantly changing, we parameterize it to a fixed size using the spatial regularity of the environment. Given a Gaussian Mixture Model (GMM)~\cite{mclachlan2000finite},
\begin{equation}
    p(x) = \sum_{k=1}^{K} \pi_k \mathcal{N}(\mathbf{x}|\boldsymbol{\mu}_k, \boldsymbol{\Sigma}_k),
\end{equation}

\noindent our framework predicts the means $\boldsymbol{\mu}_k$, standard deviations $\boldsymbol{\sigma}_k$ (used to construct diagonal covariance matrices $\boldsymbol{\Sigma}_k$), and mixing coefficients $\pi_k$. The means represent coordinates, creating a smooth probability function over the environment that assigns retention probabilities to nodes based on spatial relationships. GMM components with a mixing weight less than a specified threshold $\lambda_w$ (see Appendix~\ref{app:hyperparams}, Table~\ref{table:params}) are overridden to have large covariances and small mixing coefficients, allowing the model to selectively activate or deactivate components of the GMM. Then, the environment is split into the same patches created by the state tokenizer, and only the highest probability nodes in each patch are retained; the number of pruned nodes per patch is proportional to the number of nodes in that patch. These post-prediction steps help to improve the expressivity of the predicted GMM without requiring a large number of components. Alternative approaches we tried for predicting over a variable-sized action space are detailed in Appendix~\ref{app:discarded}.

\vspace{4pt}
\noindent \label{sec:methodology:mdp:reward}\textbf{Reward.} The reward is calculated in two stages. First, the timestep reward is calculated as the additional free space (as a fraction of the total) revealed as a result of all successful moves taken during the current timestep, minus a penalty for the difficulty of finding a suitable next move after pruning, discouraging the model from taking overly destructive actions:
\begin{equation}
R_t = \text{gain} - \lambda_A N_a,
\end{equation}

\noindent where $N_a$ is the number of attempts to find a suitable move, averaged across all successful moves taken during the current timestep, and $\lambda_A=0.05$ is the attempt penalty scaling factor.
Finally, terminal states receive a large bonus reward,
\begin{equation}
B = \lambda_B \cdot (e^C - 1),
\end{equation}

\noindent where $C$ is the fraction of the environment that has been mapped once the terminal state is reached and $\lambda_B=240$ is the scaling factor for the terminal reward. The complete reward is then defined as $R = R_t + \mathbbm{1}_{\text{terminal}} \cdot B$. The terminal bonus grows exponentially with final explored area, strongly incentivizing complete exploration, and the inclusion of timestep rewards helps with credit assignment given the sparse reward signal (discussed further in Appendix~\ref{app:discarded}).
\section{Experiments}
\label{sec:experiment}
\noindent \textbf{Setup.}
We developed a custom 2D simulation (depicted in Fig.~\ref{fig:simulator}) to conduct all experiments. It consists of a pixel grid randomly populated with obstacles, similar to the occupancy grids originally seen in frontier-based exploration~\cite{yamauchi1997frontier}.

The simulation is designed to help us narrow the focus of this work through a set of simplifying assumptions. First, the size of the environment is fixed and known. In a practical application, this means a robot using our framework has an upper limit on the area it can explore; however, real applications requiring exploration of a large area would likely already rely on multiple robots to cover the area more efficiently. Second, while a physical robot would have sensors that enable it to detect obstacles, the agent in the simulated environment has a field of view which serves this function. The simulation also disregards most uncertainty in the robot's position and in the map of the environment. In real-world applications, such uncertainty typically results from the accumulation of error in sensor measurements and Simultaneous Localization and Mapping (SLAM) systems.

The exploration algorithm we use for experiments relies on a Rapidly-Exploring Random Tree (RRT) grown within the confines of the explored area to detect frontiers. Then, a frontier is selected based on proximity and other factors, and the robot moves towards it. This method shares similarities to other RRT-based exploration strategies~\cite{umari2017rrt}, with the key difference of maintaining a global RRT rather than local RRTs.

The RL policy is trained across $10^6$ timesteps on an Intel Core i5-14600K processor and an NVIDIA RTX 5080 GPU. Hyperparameters are listed in Appendix~\ref{app:hyperparams}, Table~\ref{table:params}. During training, the model predicts a pruning strategy as described in Section~\ref{sec:methodology:mdp:action}. The number of pruned nodes is selected such that the RRT growth rate is reduced by a fixed fraction $\lambda_f=0.96$ of its normal growth rate. Once the robot has taken a fixed number of moves ($\lambda_M$) or the RRT has been grown a fixed number of times ($\lambda_R$) (both hyperparameters; see Appendix~\ref{app:hyperparams}, Table~\ref{table:params}), the simulation is reset and obstacles are randomized.

\vspace{4pt}
\noindent \textbf{Results.} As seen in Fig.~\ref{fig:reward2}, we find that despite pruning 96\% of the nodes in the RRT, our approach preserves final exploration progress within $4.5$~percentage points (pp) of the unpruned baseline across varied environment complexities. Although we trained the model on short-horizon exploration tasks, our approach generalizes well to long-horizon exploration tasks, and even exceeds the final progress of exploration without pruning by up to approximately $2.1$~pp. Further, learned pruning consistently outperforms the random pruning baseline across all experiments, enabling exploration with far fewer nodes without sacrificing on exploration efficiency.

We additionally report the absolute mean exploration progress ($\pm$ per-environment standard deviation) in Table~\ref{table:rawdata} for each pruning strategy and environment configuration, from which the percentage-point differences in Fig.~\ref{fig:reward2} are derived. Per-environment standard deviations are large since final coverage may vary substantially with the exact layout of obstacles; however, the differences in the mean coverage for all experiments remain significant despite overlapping raw distributions, since the standard error on the computed percentage point differences (shown by the error bars in Fig.~\ref{fig:reward2}) is reduced by averaging over multiple episodes.

\begin{figure}[t!]
    \centering
    % \vspace{-0.5em}
    % \includegraphics[width=0.48\linewidth]{Figures/imagem_convertida(6).png}
    % \hfill
    % \includegraphics[width=0.48\linewidth]{Figures/imagem_convertida(7).png}
    \includegraphics[width=0.77\linewidth]{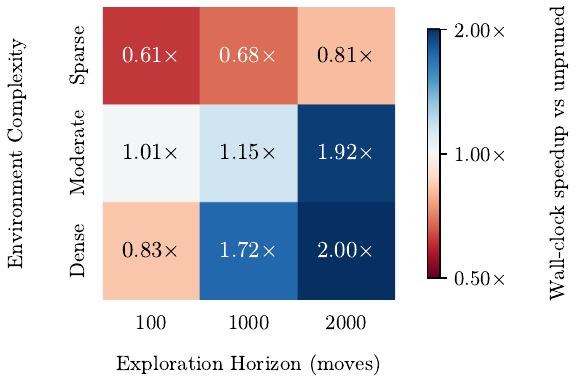}
    \vspace{-8pt}
    \caption{Heat map showing the wall-clock speedup relative to the unpruned baseline that results from applying learned pruning to various combinations of environment complexities and exploration time horizons. In sufficiently complex environments and long horizons, learned pruning achieves substantial computational efficiency gains relative to the unpruned baseline, consistent with the expected impact of a large reduction in the exploration graph size.}
    \label{fig:wallclock}
    \vspace{-8pt}
\end{figure}

Finally, we evaluate the impact of sparsification on the computational efficiency of exploration. Fig.~\ref{fig:wallclock} shows the wall-clock speedup relative to the unpruned baseline that results from applying learned pruning to various combinations of environment complexities and exploration time horizons. Especially in more complex environments and longer time horizons, learned pruning shows substantial improvements to computational efficiency, achieving up to a $2\times$ speedup, which is consistent with the expected impact of the reduction in the size of the exploration graph. In these environments, the unpruned tree accumulates redundant nodes much more rapidly, and without a mechanism to smartly remove nodes that don't contribute to the tree's structure or function, exploration slows as the algorithm searches through the entire tree (approx. 57k nodes @ $\lambda_M=\lambda_R=2000$) multiple times per step. Learned pruning preemptively removes nodes that are irrelevant and likely to cause misleading pathfinding decisions, leading to both wall-clock speedups and gains in exploration progress.

\begin{table}[t!]
\centering
\setlength{\tabcolsep}{4pt}
\renewcommand{\arraystretch}{1.05}
\caption{Final explored area (\%, mean $\pm$ std) under different pruning strategies, environment complexities, and time horizons.}
\label{table:rawdata}
    \begin{tabular}{cccccc}
        \toprule
        \textbf{\makecell{Pruning\\Method}} & \textbf{\makecell{Sparse\\100 mov.}} & \textbf{\makecell{Moderate\\100 mov.}} & \textbf{\makecell{Dense\\100 mov.}} & \textbf{\makecell{Dense\\1000 mov.}} & \textbf{\makecell{Dense\\2000 mov.}} \\
        \midrule
        \textbf{None }& \makecell{$68.94$\\$\pm14.38$} & \makecell{$50.84$\\$\pm13.36$} & \makecell{$13.49$\\$\pm4.63$} & \makecell{$29.78$\\$\pm9.22$} & \makecell{$31.99$\\$\pm12.39$} \\
        \midrule
        \textbf{Random} & \makecell{$60.52$\\$\pm14.10$} & \makecell{$44.45$\\$\pm11.36$} & \makecell{$11.15$\\$\pm3.61$} & \makecell{$22.99$\\$\pm6.87$} & \makecell{$29.82$\\$\pm9.95$} \\
        \midrule
        \textbf{\makecell{Learned\\(ours)}} & \makecell{$64.64$\\$\pm14.79$} & \makecell{$48.48$\\$\pm12.76$} & \makecell{$11.95$\\$\pm3.87$} & \makecell{$31.13$\\$\pm8.65$} & \makecell{$34.15$\\$\pm11.32$} \\
        \bottomrule
    \end{tabular}
\vspace{-12pt}
\end{table}

\section{Discussion and Conclusion}
\label{sec:conclusion}
We introduce a hybrid transformer-based framework that learns to sparsify the exploration graphs of frontier-based robotic exploration algorithms, reducing graph size by up to 96\%. The data in Section~\ref{sec:experiment} provide preliminary evidence that our framework preserves exploration performance within $4.5$~pp of the unpruned baseline and, when generalizing beyond the environments presented during training to tackle long-horizon exploration tasks in highly complex environments, that our framework exceeds the unpruned baseline by up to 2.1 pp while achieving up to a 2× wall-clock speedup.

A key limitation is the lack of comparable methods; prior graph pruning work targets static graphs with supervised learning, which does not apply to the evolving graphs in this problem, and the lack of clear heuristics makes designing a handcrafted deterministic pruner very difficult. Further, we note that the expressivity of the GMM may be a bottleneck. The post-prediction steps taken to apply the action, including patch-wise pruning and GMM component deactivation, have largely helped to combat this issue; however, replacing the GMM with a more expressive probability density function is a promising direction for further improving the performance of our approach. A more detailed discussion of this is presented in Appendix~\ref{app:discarded}. Also, we continue to observe that the model's performance is extremely sensitive to hyperparameter settings, which indicates that more work may be needed to improve the robustness of our framework. Physical-robot validation remains necessary given the simulation's assumptions about localization and environment size; however, we expect the gains in compute and exploration efficiency to carry over to real-world deployments since the pruner already tolerates some imprecision through the occupancy-grid abstraction. Overall, our results suggest that learning at the subtask level can effectively preserve the robustness of algorithmic approaches, providing a novel framework that can be refined and built upon, and opening new avenues for improving performance and reducing overhead in graph-based robotic exploration.

\newpage

\bibliographystyle{unsrt}
\bibliography{references}

% Appendixes should appear before the acknowledgment.
\newpage
% \thispagestyle{empty}
% \mbox{}

\appendices
\section{Challenges}
\label{app:discarded}

Due to the complexity of the problem, much of our work did not make it into the solution presented in this paper, and is therefore not discussed elsewhere. However, we believe it is valuable to document some of the major challenges and work done to overcome these challenges.

\subsection{Graph Encoding}
\label{app:discarded:encoding}
% \textbf{Graph Encoding.}
The biggest challenge in handling dynamically expanding graphs, such as those used in robotic exploration, is that the number of nodes in the graph is constantly changing. Therefore, these graphs cannot be passed as input to a model via traditional graph encodings, such as adjacency lists, adjacency matrices, edge lists, or incidence matrices, if the model must be trained using RL. We originally tried making use of a pretrained Graph Attention Transformer (GAT)~\cite{velickovic2018GAT} encoder to create fixed-size representations of variable-sized graphs, but designing an appropriate adjunct task to train the encoder was challenging, especially given the highly variable nature of the exploration graphs and the difficulty of discerning which features were most relevant to pruning decisions. More importantly, since all pruning decisions directly influence the structure of future input graphs, it was impractical to use a pre-trained encoder when the inputs may look drastically different once training begins for the rest of the model. It would also not have been possible to train both the GAT encoder and the rest of the model simultaneously using traditional reinforcement learning techniques, since the input to the GAT encoder is of variable length, while reinforcement learning pipelines typically require all elements of the MDP to be of fixed size.

A second approach we explored was the use of Graph2Vec~\cite{narayanan2017graph2vec}, a task-agnostic method that promises an unsupervised learning approach to encoding graphs of variable sizes while preserving structural equivalence (i.e. similar graphs produce similar embeddings). However, as with the GAT encoder, this would require pre-generating graphs to train the encoder on, and an encoder trained in this manner will likely be unable to properly represent the variety of graph structures that may arise  over the course of training. Additionally, after working with this approach for some time, it became obvious that any graph encoding approaches that directly use the graph data would make it difficult to encode the environment features and the spatial relationships between the graph and the environment. Therefore, we settled on using an image to represent the exploration graph, environment features, and spatial relationships within a fixed size input, taking advantage of our simplifying assumption that the environment would always be of a fixed, known size.

\subsection{Variable-size Action Spaces}
\label{app:discarded:gmm}
% \textbf{Variable-size Action Spaces.}
Similar to the issue with graph encoding, the model cannot directly predict probabilities for each of the nodes in the graph because the graph is of a variable size. Therefore, an intermediate, fixed-size representation of the action is necessary to map to the variable-sized set of nodes. We originally intended to make use of a solution that relies on in-context reinforcement learning to match intermediate action embeddings to desired outcomes by allowing the model to experiment with actions to observe their effect on the environment, allowing the model to adapt in real-time to action spaces not seen during training~\cite{sinii2023context}. However, this requires that changes in the action space occur infrequently, allowing the model time to build context regarding available actions. In this work, any action taken immediately modifies the set of available future actions, and this lack of an experimentation phase would prevent the model from establishing context on the effects of different action embeddings. As a result, we settled on the approach of using a probability distribution over the area of the environment to use the spatial regularity to encode actions as a fixed-size representation. Internally treating the GMM parameters as the action when training the model works since the mapping between a given set of GMM parameters and the resulting set of actions is deterministic for a given model input; i.e. for a particular environment map, a particular graph, and a particular set of GMM parameters, the set of nodes that is pruned will always be the same.

\subsection{Credit Assignment and Delayed Rewards}
\label{app:discarded:credassn}
% \textbf{Credit Assignment and Delayed Rewards.}
One of the biggest challenges in enabling the model to learn an effective policy is overcoming the credit assignment problem and the delayed reward signal the model receives. The reward signal is significantly removed from the model's prediction -- the model outputs parameters used to construct a GMM, which is then used to prune the current exploration graph, and the pruned exploration graph is then used to explore the environment further. However, the metric of performance is primarily the final coverage of the environment. So, for example, it does not matter how much coverage is gained from one exploration step to the next, as long as the environment is completely explored in the allotted time. Therefore, since the reward is significantly delayed and represents the accumulation of all actions taken over an episode, it is difficult for the model to determine which steps specifically led to the desired outcome. Combined with the fact that the state is a very information-rich but compact representation, this makes it very difficult for the model to learn an effective policy. We were able to partially remedy this issue by assigning small intermediate rewards based on the immediate benefit of pruning decisions in terms of additional knowledge gained. While this does not fully solve the issue, it was critical in enabling the model to learn a value function. However, this could also prevent the model from exploring all possible policies, such as those that see large gains in knowledge intermittently rather than cumulatively smaller but consistent knowledge gains. Further work is needed to determine the best approach to tackle this challenge.

\subsection{GMM Expressivity}
\label{app:discarded:expressivity}
Our chosen solution to combating the aforementioned problems with handling variable-sized action spaces, the use of a GMM to parameterize an action space of any size, is itself problematic in that the resulting GMM lacks significant expressivity without utilizing an unreasonably large number of component distributions.

Previously, we had noted that uniform random pruning exhibited similar raw performance to exploration without pruning under short-horizon exploration in sparsely cluttered environments, and hypothesized that uniform random pruning may serve as a component of a more effective solution. We tested this idea by introducing noise to the GMM using trigonometric functions scaled to $10^{-3}$ times the maximum GMM probability across the environment and enabling the model to selectively activate or deactivate GMM components. In regions where GMM probabilities approach zero, the noise became significant, causing behavior akin to random pruning. By activating or deactivating components, the model could control whether each region used learned or random pruning. Following this approach, we observed much greater initial exploration performance from the untrained model.

Our current solution adopts the component activation/deactivation mechanism from this experiment, along with patch-wise pruning and a slightly increased number of GMM components, which we did not originally test, and results in much greater performance overall. Regardless, this line of inquiry suggests that performance can be improved by replacing the GMM with a different probability density function to allow for greater variation with fewer parameters, since both changes have the effect of increasing the variation in pruning strategies that can be represented with a limited number of GMM components.

\newpage
\section{Hyperparameters}
\label{app:hyperparams}
% For reference, we include the following table of hyperparameter settings used during training.

\begin{table}[H]
\centering
\setlength{\tabcolsep}{4pt}
\renewcommand{\arraystretch}{1.05}
\caption{Training hyperparameters for PPO, policy networks, and simulation environment.}
\label{table:params}
    \begin{tabular}{clr}
        \toprule
        & \textbf{Parameter} & \textbf{Value} \\
        \midrule
        \multirow{11}{*}{\rotatebox{90}{\textbf{PPO}}}
        & Learning Rate ($\alpha$) & $1{\times}10^{-4}$ \\
        & Discount Factor ($\gamma$) & $0.99$ \\
        & GAE Estimation ($\lambda$) & $0.95$ \\
        & Clip Parameter ($\epsilon$) & $0.1$ \\
        & Value Function Coeff.\ & $0.5$ \\
        & Entropy Coeff.\ & $5{\times}10^{-4}$ \\
        & Update Frequency & $2048$ \\
        & K-epochs & $4$ \\
        & Num.\ Minibatch & $16$ \\
        & Target KL Divergence & $0.08$ \\
        & Warmup Episodes & $15$ \\
        \midrule
        \multirow{11}{*}{\rotatebox{90}{\textbf{Policy}}}
        & GTrXL Layer Size & $1024$ \\
        & GTrXL Layers & $3$ \\
        & Attn.\ Heads & $8$ \\
        & Attn.\ Head Size & $512$ \\
        & PwFF Size & $512$ \\
        & GTrXL Mem.\ Len.\ & $100$ \\
        & Num.\ GMM Components & $32$ \\
        & Mixing Weight Threshold ($\lambda_w$) & $0.5$ \\
        & Activation Function & $\tanh$ \\
        & Hidden Layers (Actor) & $[6400, 1600, 512, 512, 512, 512]$ \\
        & Hidden Layers (Critic) & $[6400, 1600, 512, 512, 512, 512]$ \\
        \midrule
        \multirow{4}{*}{\rotatebox{90}{\textbf{Sim.}}}
        & Env.\ Dimensions & $250$px $\times$ $250$px \\
        & Max.\ Robot Moves ($\lambda_M$) & $100$ \\
        & Max.\ RRT Growth Attempts ($\lambda_R$) & $100$ \\
        & Obstacle Density & Dense \\
        \bottomrule
    \end{tabular}
\par\vspace{8pt}
All hyperparameters listed are the settings used when training the model; several of the simulation hyperparameters specifically were varied as part of the experiments presented in the paper.
\end{table}

\section*{Acknowledgments}
Part of this work was conducted while the authors were affiliated with the Fluidic City Lab and the Tickle College of Engineering at the University of Tennessee Knoxville. We thank the Fluidic City Lab, the Tickle College of Engineering, and NVIDIA for their support of this research.

\end{document}